\documentclass{article}

\usepackage{arxiv}

\usepackage[english]{babel}
\usepackage[utf8]{inputenc} 
\usepackage[T1]{fontenc}    
\usepackage{hyperref}       
\usepackage{url}            
\usepackage{booktabs}       
\usepackage{amsfonts}       
\usepackage{microtype}      
\usepackage{lipsum}		
\usepackage{graphicx}
\usepackage[numbers]{natbib}
\usepackage{caption}
\usepackage{doi}

\usepackage{amsmath}
\usepackage{amsfonts}
\usepackage{graphicx}
\usepackage{makecell}
\usepackage{colortbl}
\usepackage[font=normalsize]{subcaption}
\usepackage{multicol}
\usepackage{multirow}
\usepackage{hhline}
\usepackage{amsmath}

\begin{document}

\title{Bayesian calibration of stochastic agent based model via random forest}

\newif\ifuniqueAffiliation
\uniqueAffiliationtrue

\ifuniqueAffiliation 
\author{ \href{https://orcid.org/0000-0003-1762-3469}{\includegraphics[scale=0.06]{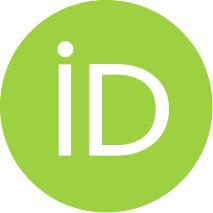}\hspace{1mm}Connor Robertson}\\
	Sandia National Laboratories\\
	Livermore, CA \\
	\texttt{cjrobe@sandia.gov} \\
	\And
    Cosmin Safta \\
	Sandia National Laboratories\\
	Livermore, CA  \\
	\And
    Nicholson Collier \\
	Argonne National Laboratory\\
	Chicago, IL \\
	\And
    Jonathan Ozik \\
	Argonne National Laboratory\\
	Chicago, IL  \\
	\And
    Jaideep Ray \\
	Sandia National Laboratories\\
	Livermore, CA  \\
}

\maketitle

\begin{abstract}
Agent-based models (ABM) provide an excellent framework for modeling outbreaks and interventions in epidemiology by explicitly accounting for diverse individual interactions and environments. However, these models are usually stochastic and highly parametrized, requiring precise calibration for predictive performance. When considering realistic numbers of agents and properly accounting for stochasticity, this high dimensional calibration can be computationally prohibitive. This paper presents a random forest based surrogate modeling technique to accelerate the evaluation of ABMs and demonstrates its use to calibrate an epidemiological ABM named CityCOVID via Markov chain Monte Carlo (MCMC). The technique is first outlined in the context of CityCOVID's quantities of interest, namely hospitalizations and deaths, by exploring dimensionality reduction via temporal decomposition with principal component analysis (PCA) and via sensitivity analysis. The calibration problem is then presented and samples are generated to best match COVID-19 hospitalization and death numbers in Chicago from March to June in 2020. These results are compared with previous approximate Bayesian calibration (IMABC) results and their predictive performance is analyzed showing improved performance with a reduction in computation.
\end{abstract}


\renewcommand\thefootnote{}

\renewcommand\thefootnote{\fnsymbol{footnote}}
\setcounter{footnote}{1}

\section{Introduction}

Agent-based models (ABMs) are powerful tools for simulating complex systems that have found use across diverse domains, from traffic flow and ecology to economics and epidemiology.
These "bottom-up" computational frameworks represent systems as a collection of autonomous agents that interact with each other and with their environment.
This decentralized, microscopic perspective allows ABMs to capture small scale or emergent phenomena that traditional ``top-down,'' or population level, approaches often miss.

ABMs find applications in a plethora of fields.
In traffic flow, they have been used to identify transportation bottlenecks and explore conditions that reduce the efficiency of infrastructure~\cite{hofer2018traffic,arnaout2010traffic,ljubovic2009traffic,shaharuddin2023traffic}.
In ecology, they can accurately model the spread of invasive species and predict ecosystem tipping points~\cite{jiang2018ecology,keith2013ecology,rebaudo2011ecology}.
In economics, they simulate market dynamics and assess the impact of policy interventions~\cite{dawid2008economics,axtell2022economics,vermeulen2016economics}.
Despite their versatility, ABMs have a fundamental challenge: parameter calibration. While ABMs have been shown to effectively replicate historical data and trends, they often include a wide range of possible individual and environmental characteristics making calibrating ABMs a high-dimensional problem.
Further, ABM simulations often scale poorly due to agent interactions, making each run of the model computationally expensive.
To compound these challenges, ABM models are often inherently stochastic.
Unlike deterministic models, ABMs incorporate randomness in agent behaviors and in environmental factors.
Though this stochasticity is crucial for capturing real-world complexity, it introduces significant uncertainty, making precise calibration elusive and increasing the computational expense of calibration.

Various calibration techniques for ABMs have been proposed to align simulations with empirical data.
These include Approximate Bayesian Computation (ABC)~\cite{de2023abm_abc,van2015abm_abc,rutter2019imabc}, variational inference~\cite{dong2016abm_vi,lenti2024abm_vi}, Markov chain Monte Carlo (MCMC)~\cite{grazzini2017abm_mcmc_abc}, and evolutionary algorithms~\cite{moya2021abm_evolutionary,robles2021abm_evolutionary}.
These parameter estimation approaches are commonly combined with an emulator or surrogate model of the ABM such as Gaussian processes~\cite{de2015abm_surrogates_gp,ahmed2023abm_surrogates_gp,fadikar2018abm_surrogates_gp_calibration,ozik2021population}, decision trees (or forests)~\cite{kieu2024abm_surrogates_rf,angione2022abm_surrogates_rf,lamperti2018abm_surrogates_rf,ozik_extreme-scale_2018}, or ordinary differential equations (ODEs)~\cite{avegliano2023abm_surrogates_ode} to reduce computational cost.
However, each of these existing calibration approaches face limitations, including their ability to address stochasticity in the underlying ABM, guarantees on their convergence, and their computational efficiency.

This paper proposes a novel approach to calibrating ABMs by introducing a random forest based global surrogate model which can connect the nonlinear dependence of population level outputs to the ABM parameters over long temporal stretches.
This approach includes decomposing quantities of interest via principal component analysis (PCA) and using the built-in sensitivity measures of the random forest to reduce the dimensionality.
This surrogate is combined with Bayesian sampling, in the form of MCMC, to produce approximate posterior distributions for the ABM parameters of interest in a fraction of the time it would take using repeated ABM evaluations.
Rigorous validation metrics are used to quantify the success of the calibration and the resulting posterior distributions are sampled to produce an ABM generated ``pushforward'' comparison.
Though generally applicable to any ABM calibrated to population-level observations, we will present the calibration approach in the context of the epidemiological ABM CityCOVID~\cite{ozik2021population,hotton_impact_2022}, which was used to model the spread of COVID-19 in the greater Chicago area during 2020 and supported city and state public health decision making.

The remainder of the paper is structured as follows.
Section 2 outlines the CityCOVID ABM, its use, and respective calibration and surrogate training data.
Section 3 outlines the surrogate construction procedure and details the formulation of the calibration problem for this ABM.
Sections 4 and 5 discuss the results of the calibration procedure for CityCOVID and conclude the paper with key findings, future directions, and the broader implications of our work for advancing ABM calibration.

\subsection{Literature review}
\label{sec:litrev}

There are a variety of approaches which have been explored to calibrate epidemiological ABMs.
Among these is the approach by Fadikar et al.~\cite{fadikar2018abm_surrogates_gp_calibration} who used Gaussian process surrogates to model the mean evolution of an epidemiological ABM and its quantile evolution to capture stochasticity.
Further, Anirudh et al.~\cite{anirudh2022abm_surrogates_nn_calibration} presented a surrogate modeling approach which first decomposed data into temporal modes using PCA and then modeled the mapping from ABM parameters to PCA weights using a neural network.
In each of these cases, the parameter estimation was performed with either rejection ABC or MCMC.

Calibration without the use of surrogate accelerators has been attempted by way of genetic algorithms, which can identify successful parameter values in search spaces of moderate dimension but they do not provide uncertainty quantification~\cite{calvez2005calibration_genetic}.
This approach allows for global exploration of the parameter space of ABMs but requires specifying fitness functions which are often problem specific and do not have any convergence guarantees.
Given the stochasticity of most ABMs and the often limited data used for calibration, having reliable approaches that include estimated uncertainty on the parameter calibration is a necessity.

In the last decade, work has focused on leveraging more advanced machine learning approaches to improve calibration or to accelerate calibration with improved surrogate models.
For example, LightGBM gradient boosted forests of decision trees have been used to filter proposed parameters in ABC sampling~\cite{panovska2023machine}.
Alternatively, causal structure has been prescribed to fit surrogates made up of systems of ordinary differential equations or recurrent neural networks and capture ABM outputs in a latent space~\cite{dyer2023interventionally}.
As a summary of surrogate approaches, Angione et al.~\cite{angione2022using} provided a comparison of various common machine learning algorithms applied as surrogate models for an example ABM.
In this, neural networks were shown to be the most effective at replicating ABM outputs, but random forests were identified as a close competitor and far more computationally efficient.
Additionally, the nonlinearity and stochasticity of the mapping from ABM parameters to outputs was cited as a key challenge and PCA decomposition identified as a potential aid for surrogate accuracy over timeseries outputs.
Across previous work, the stochasticity of ABMs was incorporated into surrogate modeling by including the random seed as an input parameter, reducing the individual seeded runs to deterministic models~\cite{panovska2023machine,dyer2023interventionally,angione2022using}.

The most recent work for ABM calibration has begun exploring the possibility of differentiable ABMs, which can automatically provide information for calibration.
New frameworks for ABM construction built on computational tools for deep learning can provide gradients for optimization automatically at evaluation of ABMs~\cite{chopra2022differentiable}.
Fundamentally, this new approach can be extended even through discrete randomness, a characteristic feature of most ABMs~\cite{arya2022automatic}.
These new approaches hold the potential of accelerating the calibration procedure using Hamiltonian-based MCMC approaches, which require gradients of the posterior distributions with respect to the parameters.
This opens the possibility of calibrating high dimensional systems while maintaining the convergence guarantees of MCMC approaches.
However, incorporating these automatic gradient computations requires fundamental reconstruction of the software for ABM modeling.
The approach presented in this paper instead takes a black-box perspective which requires no intrusive modification of the model.

Properly assessing the accuracy of stochastic model calibration is often fraught with nuance and delicacy.
However, one effective approach is the use of strictly proper scoring metrics~\cite{gneiting2007crps} which allow for unique maximizers while still providing the flexibility to match the problem at hand.
Most common among these metrics is the continuous ranked probability score (CRPS~\cite{05gr4a}) which generalizes mean absolute error for predictive cumulative distribution functions where a CRPS of 0 is a perfect match of an ensemble of stochastic simulations and the observation and positive values represent mean absolute error between each ensemble run and the observation.
This scoring technique has been successfully used to evaluate the efficacy of ABMs previously~\cite{wang2020crps_application}.
Further validation can be found by considering the verification rank histogram~\cite{hamill2001rank_verification_hist} (VRH) of the ABM outputs which can identify over or under-dispersion or equivalently tendencies to over- or under-predict compared to observational data.

\section{CityCOVID}
\label{sec:abm}

CityCOVID is an ABM developed during the COVID-19 pandemic to quantify the outcomes of behavioral and policy interventions in the Chicago, IL metropolitan area.
Based on the epidemiological ABM framework ChiSIM~\cite{macal2018chisim} and the Repast HPC distributed ABM toolkit~\cite{collier_parallel_2013}, CityCOVID includes an age-stratified population of 2.7 million agents which occupy 1.2 million distinct locations including households, workplaces, schools, nursing homes, hospitals, and gyms.
The agents move between these locations according to a variety of schedules based on their demographics~\cite{timeuse} and can be exposed to infection when present at a location with infected agents.
Individual agents transition between epidemiological states: susceptible, exposed, presymptomatic, infected (asymptomatic), infected (symptomatic), hospitalized, hospitalized (ICU), recovered, and deceased.
The transitions are governed by probabilities of exposure to infected, hospitalization, or death and the durations of agent infections and hospitalizations are Gamma distributed.

In order to capture the heterogeneity and complexity of epidemiological dynamics between individuals, agents and their interactions are governed by draws from parametrized probability distributions.
This high-dimensional parametrization gives flexible control over the impacts of different policies and interventions but also yields a wide range of possible outcomes and introduces enormous stochasticity, making the model a challenge to calibrate.

Given this challenge and its dimensionality, full Bayesian calibration is computationally infeasible.
However, previous work by Ozik et al.\cite{ozik2021population} used a sequential approximate Bayesian calibration (IMABC) approach to iteratively sample from a prior distribution which evolves over time\cite{rutter2019imabc}.
At each iteration, this distribution is updated by comparing the averaged hospitalization and death trajectories of stochastic realizations of  CityCOVID parameter combinations with the empirical data observed in Chicago during 2020.
This approach requires a large number of runs but is more efficient than comparable rejection ABC methods.
Its efficiency was further improved by performing a Morris global sensitivity analysis~\cite{morris1991sensitivity}, which allowed for a reduction to only 9 CityCOVID parameters that strongly influence the output hospitalization and death trajectories of the model.
These parameters are listed in Table~\ref{tab:citycovid-details}.

Even with parameter reduction and IMABC optimization, the model calibration required just over 32,000 runs which amounted to a total of 420,000 core hours on the Argonne Leadership Computing Facility Theta supercomputer.
Relevant figures and details on CityCOVID and its calibration can be found in Ozik et al.\cite{ozik2021population} and specifics on the IMABC algorithm in Rutter et al.\cite{rutter2019imabc}.
The mean outcomes of the calibrated model were a good match for the data and the predictive accuracy of these outcomes was sufficient to use as forecasting support for city and state public health response.
This paper furthers this work by targeting a reduction in the computational burden of the calibration while also reducing the uncertainty in the posterior estimations.

\begin{table}
    \centering
    \begin{multicols}{2}
        \begin{tabular}{|m{.3\textwidth}|}
        \hline
        \rowcolor[gray]{0.8}
        \textbf{Rates} \\
        \hline
        * Exposure to infected \\
        \hline
        \end{tabular}

         \vspace{.3cm}

        \begin{tabular}{|m{.3\textwidth}|}
        \hline
        \rowcolor[gray]{0.8}
        \textbf{Probabilities} \\
        \hline
        * Stay at home \\
        * Protective behaviors \\
        \hline
        \end{tabular}

         \vspace{.3cm}

        \begin{tabular}{|m{.3\textwidth}|}
        \hline
        \rowcolor[gray]{0.8}
        \textbf{Proportions} \\
        \hline
        Isolating in home \\
        Isolating in nursing home \\
        \hline
        \end{tabular}

        \begin{tabular}{|m{.3\textwidth}|}
        \hline
        \rowcolor[gray]{0.8}
        \textbf{Multipliers} \\
        \hline
        Seasonality \\
        \hline
        \end{tabular}

        \vspace{.3cm}

        \begin{tabular}{|m{.3\textwidth}|}
        \hline
        \rowcolor[gray]{0.8}
        \textbf{Other} \\
        \hline
        * Time of initial seeding of infections \\
        Number of initially infected \\
        Shielding by other susceptible \\
        \hline
        \end{tabular}

    \end{multicols}
    \caption{The most influential parameters on the census outputs of hospitalizations and deaths in the CityCOVID ABM based on the Morris global sensitivity analysis. Parameters marked with stars represent those most influential to the surrogate model.}
    \label{tab:citycovid-details}
\end{table}

\subsection{Data}
\label{sec:data}

In this section we describe the observational data used in model calibration as well as the process of generating data suitable for use in training a surrogate model to approximate CityCOVID.

\noindent{ {\bf Observational data: }}
The daily census numbers of occupied hospitalization beds and cumulative deaths caused by COVID-19 were collected by the Illinois National Electronic Disease Surveillance System in Chicago from March to June of 2020.
Due to the lack of COVID-19 testing capability early in 2020, case counts of COVID-19 were unreliable for use with calibration.

\noindent{ {\bf Surrogate training dataset: }}
In order to accurately reproduce CityCOVID results with a surrogate model, a representative sample of the model outputs in a parameter range of interest was needed.
The IMABC calibration effort (see \S\ref{sec:abm}) provided a 9 dimensional parameter space on which the model generates realistic rates of hospitalizations and deaths.
By first training a surrogate model (described in \S~\ref{sec:surrogate}) on these previous simulations, feature importance could be recalculated to narrow the parameter space down to the top 4 parameters using the sensitivity metrics of the surrogate (random forest~\cite{huang2016rf_sensitivity}).
The sensitivity of the parameters for the surrogate are listed in Table~\ref{tab:rf_sensitivity} and the 4 retained parameters ($\vec{\theta}$) are marked with stars in Table~\ref{tab:citycovid-details}.
The prior beliefs for these parameters were chosen to be non-informative within a region empirically chosen by comparison of the IMABC calibration outputs and the observational dala and are shown in Table~\ref{tab:priors}.

\begin{table}
    \centering
    \begin{tabular}{|c|l|c|}
        \hline
        \rowcolor[gray]{0.8}
        \textbf{Parameter} &
        \textbf{Description} &
        \textbf{Prior Belief} \\ \hline
        $\theta_1$ &
        Rate of exposure to infected &
        $\mathcal{U}(0.046,0.069)$ \\ \hline
        $\theta_2$ &
        Time of initial seeding of infections &
        $\mathcal{U}(31,59)$ \\ \hline
        $\theta_3$ &
        Probability of stay at home &
        $\mathcal{U}(0.939,0.981)$ \\ \hline
        $\theta_4$ &
        Probability of protective behaviors &
        $\mathcal{U}(0.407,0.492)$ \\ \hline
    \end{tabular}
    \caption{Prior beliefs for 4 retained parameters for surrogate training and calibration. The range of the uniform distributions was taken from preliminary runs of CityCOVID~\cite{ozik2021population}.}
    \label{tab:priors}
\end{table}

To give the surrogate adequate training information in these prior domains, 700 quasi-random samples of $\vec{\theta}$ were taken using Halton sampling\cite{wong1997haltonsampling} in the 4 dimensional space. Each of the 700 parameter sets were simulated with 50 different random seeds, allowing for a robust characterization of stochasticity to be represented in the outcome. These seeds alter the selection of who is initially infected in the simulation, random draws related to infectious state duration and transmission, and in location movements within agent schedules. The resulting hospitalization and death projections for each simulation were then averaged across random seeds to produce a ``mean-model'' estimate of the parameter outputs.

This approach does not fully account for the stochasticity of the ABM but reflects the most common approach used for ABM-based forecasting. Comparisons of the complete dataset with the observed hospitalizations ($\hat{h}$) and deaths ($\hat{d}$) are shown in Figure~\ref{fig:all_abm_data}.
Note that although the hospitalization and death information in CityCOVID is tied to specific agents and thus spatially distributed, here we only consider comparisons of the Chicago city-wide census quantities.

\begin{figure}
    \centering
    \includegraphics[width=\textwidth]{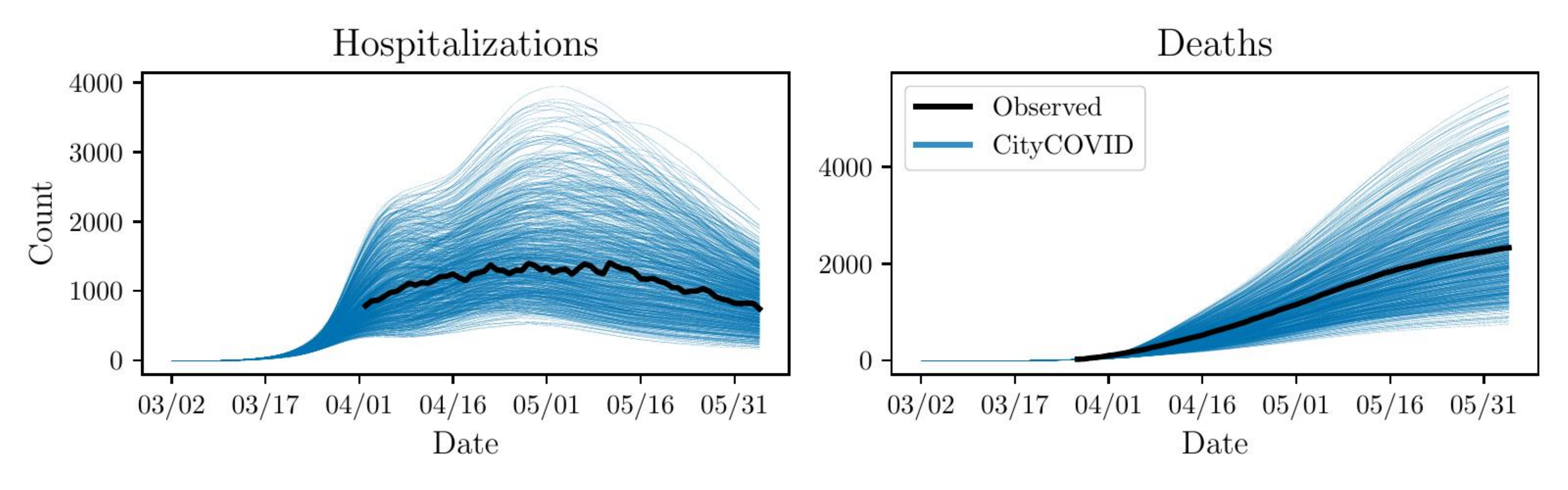}
    \caption{Range of hospitalization and death trajectories for the observed data in Chicago from March-June of 2020 (black) and the CityCOVID simulations using parameter values from the 4 dimensional quasi-random hypercube used to train and test the surrogate method (blue). CityCOVID outputs are averaged across random seeds.}
    \label{fig:all_abm_data}
\end{figure}

\section{Methods}

\subsection{Surrogate model} \label{sec:surrogate}

Prior work with ABMs has demonstrated the highly nonlinear nature of their dynamics~\cite{manson2012abm_complexity} and as such, reductions which characterize the temporal behavior into smooth modes have been effective~\cite{fadikar2018abm_surrogates_gp_calibration, anirudh2022abm_surrogates_nn_calibration}.
Following these approaches, this surrogate approach begins by decomposing the temporally concatenated hospitalization and death series using PCA:

\begin{align}
    \begin{bmatrix}
        h_{1,1} & \ldots & h_{1,n} & d_{1,1} & \ldots & d_{1,n} \\
        h_{2,1} & \ldots & h_{2,n} & d_{2,1} & \ldots & d_{2,n} \\
        \vdots & \ldots & \vdots & \vdots & \ldots & \vdots \\
        h_{m,1} & \ldots & h_{m,n} & d_{m,1} & \ldots & d_{m,n}
    \end{bmatrix}
    \rightarrow
    \begin{bmatrix}
        \alpha_{1,1} \\ \alpha_{2,1} \\ \vdots \\ \alpha_{m,1}
    \end{bmatrix}\odot
    \vec{c}_1
    + \ldots +
    \begin{bmatrix}
        \alpha_{1,2n} \\ \alpha_{2,2n} \\ \vdots \\ \alpha_{m,2n}
    \end{bmatrix}\odot
    \vec{c}_{2n}
\end{align}
\noindent where $h_{i,j},d_{i,j}$ are hospitalizations and deaths for parameter set $i$ at time step $j$, $\vec{c}_j \in \mathbb{R}^{2n}$ are the PCA components, and $\alpha_{i,j}$ are the coefficients multiplying components $\vec{c}_j$ for parameter set $i$.
To reduce the dimensionality of the surrogate mapping from ABM parameters to component coefficients, the PCA decomposition is truncated to only the most dominant components $\vec{c}_j$ in order to capture 95\% of the temporal variance.

Given this decomposition, a regressor $R^s_{\vec{\gamma}}$ is trained to map the Morris sensitivity-reduced CityCOVID ABM parameter set (9 parameters) to the output coefficients $\vec{\alpha}_i = \{\alpha_{i,j}\}$ for all time steps $j= 1,\ldots,n$ with regressor hyperparameters $\vec{\gamma}$.
Although a wide range of classical or machine learning surrogate approaches could be used for regressor $R^s_{\vec{\gamma}}$~\cite{anirudh2022abm_surrogates_nn_calibration}, random forests were selected due to their computational efficiency and because their structure naturally capture both nonlinear and discontinuous behaviors, both of which are characteristic in ABMs~\cite{manson2012abm_complexity}.

To improve the computational efficiency of the surrogate-based calibration, the dimensionality of the ABM parameter set was reduced by analyzing the sensitivity of the random forest with Gini impurity\cite{hastie2009gini}, permutation importance\cite{hastie2009gini}, and Sobol indices\cite{sobol2001global}.
These sensitivity metrics measure how important each input parameter is to: the structure of the trees (Gini), the scale of the outputs (permutation), and the variance of the output (Sobol).
They are discussed in detail in Appendix~\ref{asec:surrogate_performance} and the exact sensitivity values are listed in Table~\ref{tab:feature_importance}.
As can be observed, the rate of exposure to infected, i.e., the hourly probability of getting exposed from an infected individual, and the time of initial seeding of infections play an outsized role on the hospitalization and death trajectories, according to the random forest.

\begin{table}
\centering
\begin{tabular}{|l|c|c|c|c|}
\hline
\rowcolor[gray]{0.8}
\textbf{Feature} &
\textbf{Gini Importance} &
\textbf{Permutation Importance} &
\textbf{Sobol (first)} &
\textbf{Sobol (total)}
\\ \hline
Rate of exposure to infected &
0.41 &
0.59 &
0.52 &
0.56
\\ \hline
Time of initial seeding of infections &
0.32 &
0.61 &
0.29 &
0.33
\\ \hline
Probability of stay at home &
0.18 &
0.29 &
0.08 &
0.09
\\ \hline
Probability of protective behaviors &
0.08 &
0.05 &
0.01 &
0.01
\\ \hline
\end{tabular}
\caption{Random forest feature importance metrics for CityCOVID parameters}
\label{tab:feature_importance}
\end{table}

After reducing the input dimensionality, a final random forest $R_{\vec{\gamma}}$ was trained to map only the surrogate sensitivity-reduced CityCOVID ABM parameter set $\vec{\theta}_i = \{\theta_{1}, \theta_{2}, \theta_{3}, \theta_{4}\}_i$ to the output coefficients $\vec{\alpha}_i = \{\alpha_{i,j}\}$.
This random forest $R_{\vec{\gamma}}$ has several hyperparameters $\vec{\gamma}$ which can dramatically affect its performance including the number of trees, the criterion used to split trees, and constraints on the number of samples needed for splits and leaves.
In order to maximize accuracy, the hyperparameters were tuned via 5-fold cross validation brute-force search.
Given the optimally selected hyperparameters, the random forest is trained with the full hypercube of surrogate sensitivity-reduced data.

\subsection{Formulation of the estimation problem}
\label{sec:calibration}

Bayesian methods are a desirable approach for calibration due to their control over posterior form and convergence guarantees but are only practical for low dimensional calibrations.
Our problem easily fit this requirement after reduction of the parameter space using surrogate sensitivity.

Though the surrogate model was created to estimate the census trajectories of hospitalizations and deaths, the non-stationarity of these features is prohibitive for convergence of Bayesian sampling via MCMC.
This is because the census values of hospitalizations and deaths are of different scales and fluctuate over large ranges, the likelihood calculation can be biased toward approximating death curves for later dates.
As a result, calibration is performed using rolling averages over 1-week windows for the daily counts of hospitalizations ($\vec{h}^\circ$) and deaths ($\vec{d}^\circ$) which are computed via finite differences. The surrogate outputs were also translated into daily counts by forward finite differences.

Let $(\vec{h}, \vec{d}) = \mathcal{M}(\vec{\theta})$ be the daily predictions of the CityCOVID model, conditional on input parameters $\vec{\theta} = \{\theta_1, \theta_2, \theta_3, \theta_4\}$, as defined in Table~\ref{tab:priors}. Here $\vec{h} = \{h_j\}$ and $\vec{d} = \{d_j\}, j = 1 \ldots n$ are the daily counts of hospitalizations and deaths produced by the model. Let $\vec{h}^{\circ}$ and $\vec{d}^{\circ}$ be their observed counterparts linked by a zero-mean Gaussian error. I.e.,
\[
    h_j^{\circ} = h_j(\vec{\theta}) + \epsilon_h, \epsilon_h \sim \mathcal{N}(0, \sigma_h^2)
    \mbox{\hspace{3mm} and \hspace{3mm}}
    d_j^{\circ} = d_j(\vec{\theta}) + \epsilon_d, \epsilon_d \sim \mathcal{N}(0, \sigma_d^2).
\]
The likelihood of observing $(\vec{h}^{\circ}, \vec{d}^{\circ})$, conditional on $\vec{\theta}$, is
\begin{eqnarray*}
    \mathcal{L}(\vec{h}^{\circ}, \vec{d}^{\circ} \mid \vec{\theta}) & = & \frac{1}{(2 \pi)^{n/2} \sigma_h^n} \prod_{j = 1}^n \exp{\left[ - \frac{1}{2} \left( \frac{h_j^{\circ} - h_j(\vec{\theta})}{\sigma_h}\right)^2\right]} \times \frac{1}{(2 \pi)^{n/2} \sigma_d^n} \prod_{j = 1}^n \exp{\left[ - \frac{1}{2} \left( \frac{d_j^{\circ} - d_j(\vec{\theta})}{\sigma_d}\right)^2\right]} \nonumber \\
    & = & \frac{1}{(2\pi \sigma_d \sigma_h)^n} \exp{\left[ - \frac{S_h}{2 \sigma_h^2} - \frac{S_d}{2 \sigma_d^2}\right]},
\end{eqnarray*}
where $S_h = \sum_{j = 1}^n \left(h_j^{\circ} - h_j(\vec{\theta}) \right)^2$, $S_d = \sum_{j = 1}^n \left(d_j^{\circ} - d_j(\vec{\theta}) \right)^2$ and $\left(h_j(\vec{\theta}), d_j(\vec{\theta}) \right)$ are the model predictions corresponding to parameters $\vec{\theta}$.

Let $\pi (\vec{\theta})$ be the prior belief of $\vec{\theta}$ as listed in Table~\ref{tab:priors} i.e., $\vec{\theta} \sim \mathcal{U}\left(\vec{\theta}^{l}, \vec{\theta}^u \right)$, where $\left(\vec{\theta}^l, \vec{\theta}^u \right)$ are the lower and upper bounds in Table~\ref{tab:priors}. The error variances $(\sigma_h^2, \sigma_d^2)$ are modeled with conjugate priors, i.e., with an inverse Gamma prior or,
\[ \sigma_h^{-2} \sim \mathcal{G}\left(\frac{n_s + n}{2}, \frac{n_s \zeta_h^2 + S_h}{2} \right) \mbox{\hspace{3mm} and \hspace{3mm}}
   \sigma_d^{-2} \sim \mathcal{G}\left(\frac{n_s + n}{2}, \frac{n_s \zeta_d^2 + S_d}{2} \right), \]
where $n_s \zeta_h^2 + S_h$ and $n_s \zeta_d^2 + S_d$ are the \emph{rate} parameters of the Gamma ($\mathcal{G}$) distributions and $n$ is the number of observations. This distribution leads to $\left(\zeta_h^2, \zeta_d^2 \right)$ being the prior means of $ \left(\sigma_h^2, \sigma_d^2 \right)$ and $n_s$ is a user-defined value that, together with $\left(\zeta_h^2, \zeta_d^2 \right)$, defines their prior variance.  In this paper we use $n_s = 1$ which implies a noninformative prior, $\zeta_h^2 = \frac{S_h^\text{OLS}}{n-p}, \zeta_d^2 = \frac{S_d^\text{OLS}}{n-p}$ where $S_h^\text{OLS},S_d^\text{OLS}$ are determined using the ordinary least squares (OLS) optimal parameter set from the dataset used for surrogate training, and $p = 4$ as the number of parameters in $\vec{\theta}$.

By Bayes rule, the likelihood $\mathcal{L}$ and the priors can be combined into an expression for the posterior distribution for $\vec{\theta}$,
\begin{equation}
    P \left(\vec{\theta} \mid \vec{h}^\circ, \vec{d}^\circ \right) \propto \frac{1}{(2\pi \sigma_d \sigma_h)^n} \exp{\left[ - \frac{S_h}{2 \sigma_h^2} - \frac{S_d}{2 \sigma_d^2} \right] } \times
    \sigma_h^{n_s/2 -1} \sigma_d^{n_s/2 - 1} \exp{\left( - \frac{n_s \zeta_h^2}{2} - \frac{n_s \zeta_d^2}{2}\right)} \times
    \pi\left( \vec{\theta} \right).
    \label{eq:likelihood}
\end{equation}

Delayed rejection adaptive Metropolis-Hastings sampling (DRAM)~\cite{haario2006dram} was used to draw samples from the posterior in Equation~\ref{eq:likelihood}. Strictly speaking, each step of the algorithm consists of a DRAM update of $\vec{\theta}$ followed by a Gibbs update
of $\left(\sigma_h^2, \sigma_d^2 \right)$, if the proposal for $\vec{\theta}$ is accepted by DRAM.
The implementation of DRAM is available in the \texttt{pymcmcstat} Python package\cite{19mp1a}.
Since each iteration of DRAM requires an evaluation of CityCOVID, the use of the surrogate model described in \S~\ref{sec:surrogate} makes the algorithm feasible.
DRAM yields a Markov chain of samples $\vec{\theta}_k,\ k =\{1, \ldots, K\}$ and in our study, $K = 50,000$ steps.
This sequence is checked for stationarity as a stopping criterion, using the method by Raftery and Warnes\cite{Warnes:2000}, which builds on an older method by Raftery and Lewis\cite{92rl2a}. This method is implemented as an R version 4.3.3 (2024-02-29)\cite{R:Manual} package \texttt{mcgibbsit}\cite{mcgibbsit:Manual}. The package computes the minimum run length $N_{min}$, the required burn-in $M$, and the number of samples required to meet an estimation accuracy criterion for each component of $\vec{\theta}$.

\section{Results}

\subsection{Surrogate performance}
\label{sec:surr_perf}

The reconstruction of the data from 4 PCA components yields a median absolute relative error of 2\%.
Figure~\ref{fig:pca_scree} shows a scree plot demonstrating the variance explained as the number of principal components is increased.
As can be observed, the temporal dynamics present in the hospitalizations and death trajectories from the ``mean-model'' of CityCOVID are smooth and fairly simple, allowing for efficient encoding in the principal components.
An example comparison of a CityCOVID trajectory with its respective PCA compression is shown in Figure~\ref{fig:pca_reconstruct_32}.
Note that the relative error is significantly higher at early times when the number of hospitalizations and deaths are low due to division by small numbers in the relative error calculation.
These small numbers do not affect the likelihood estimation in Equation~\ref{eq:likelihood} and are not discussed further.

\begin{figure}
    \begin{subfigure}{.49\textwidth}
        \centering
        \includegraphics[width=\textwidth]{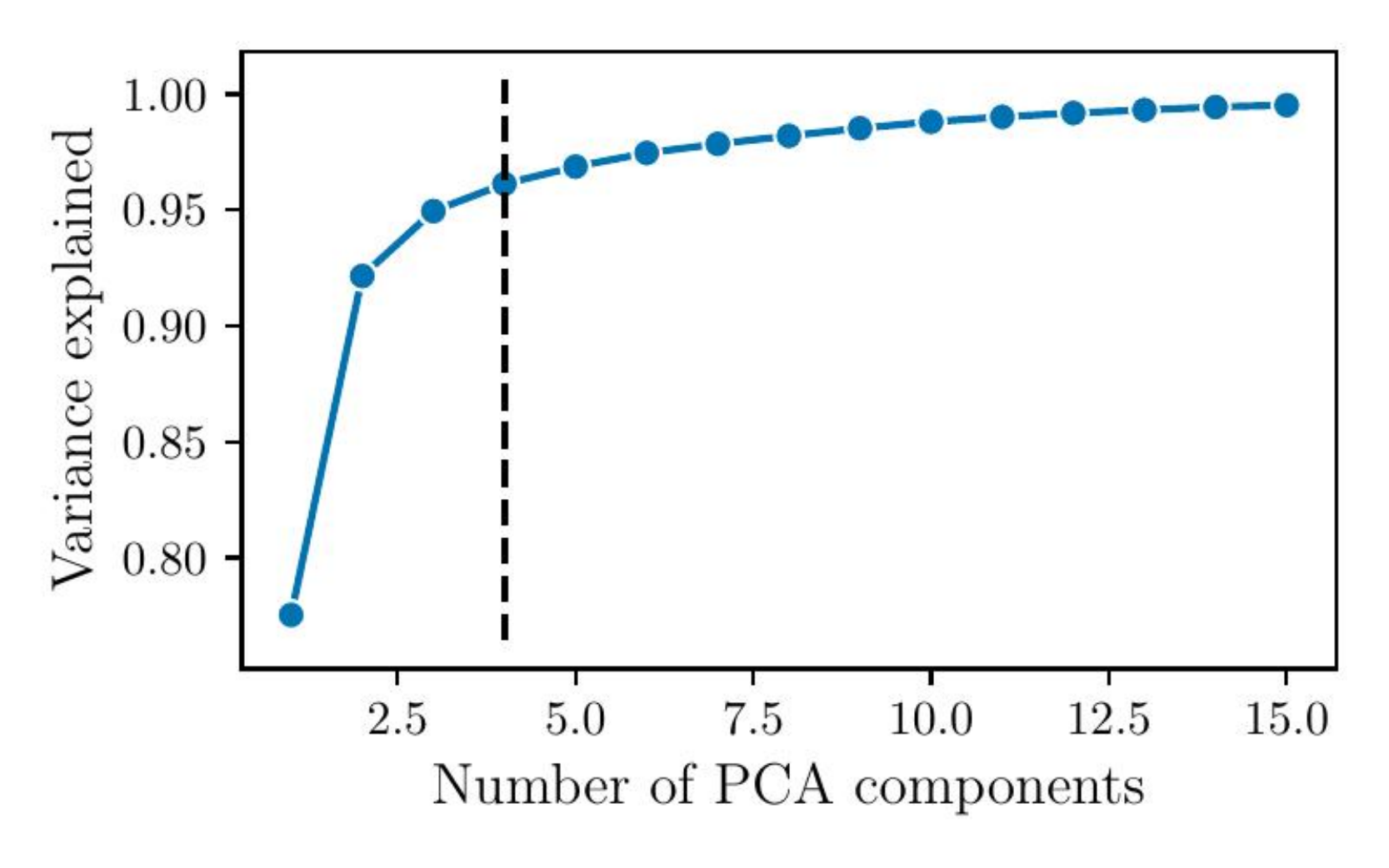}
        \caption{}
        \label{fig:pca_scree}
    \end{subfigure}
    \begin{subfigure}{.49\textwidth}
        \centering
        \includegraphics[width=\textwidth]{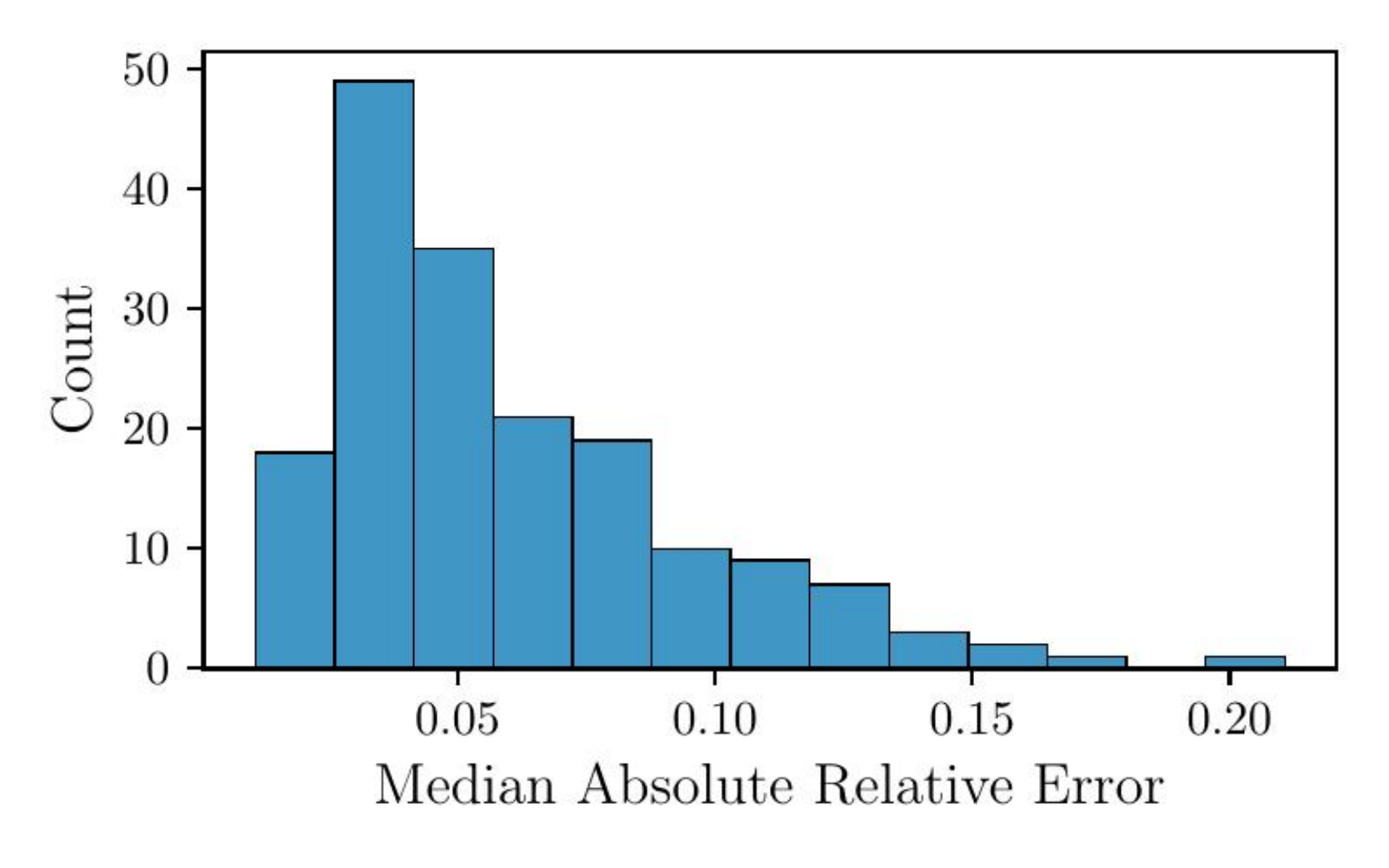}
        \caption{}
        \label{fig:rel_err_hist}
    \end{subfigure}
    \caption{(a) Scree plot demonstrating approximation power of different numbers of PCA modes with a dotted line at the number of modes used for the surrogate. (b) Median absolute relative error for surrogate reconstructions of CityCOVID hospitalization and death trajectories.}
    \label{fig:surrogate_err}
\end{figure}

\begin{figure}
    \centering
    \includegraphics[width=\textwidth]{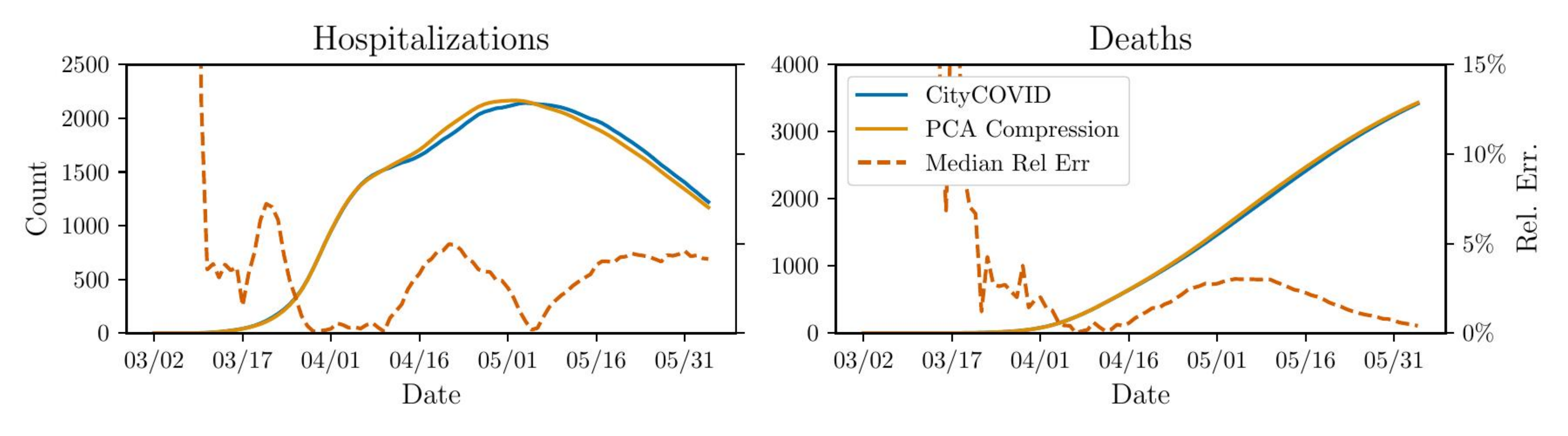}
    \caption{Accuracy of the data reconstructed using 4 principal components. The components capture over 95\% of the variance of the data.}
    \label{fig:pca_reconstruct_32}
\end{figure}

The random forest trained to reconstruct the original trajectories was able to achieve a median absolute relative error of less than 5\% over five fold cross validation.
The distribution of the median absolute relative errors for a random testing set of holdout trajectories is shown in Figure~\ref{fig:rel_err_hist} and several random examples demonstrating the surrogate predicted hospitalization and death curves as compared with CityCOVID trajectories are shown in Appendix~\ref{asec:surrogate_performance}.

\begin{table}
    \centering
    \begin{tabular}{|c|l|c|}
        \hline
        \rowcolor[gray]{0.8}
        \textbf{Hyperparameter} &
        \textbf{Description} &
        \textbf{Value} \\ \hline
        $\gamma_1$ &
        number of trees &
        500 \\ \hline
        $\gamma_2$ &
        split quality criterion &
        absolute error \\ \hline
        $\gamma_3$ &
        minimum number of samples per leaf &
        3 \\ \hline
        $\gamma_4$ &
        max number of features per split &
        5 \\ \hline
    \end{tabular}

    \caption{Descriptions and values for random forest hyperparameters after brute force search. Values were selected using 5 fold cross validation.}
    \label{tab:rf_hyperparams}
\end{table}

\subsection{Parameter estimation}
\label{sec:calib}
 Using the surrogate described in Section~\ref{sec:surrogate}, approximate hospitalization and death trajectories were sampled for use with the MCMC sampling of $P(\vec{\theta} \ | \ h,d)$ as described in Section~\ref{sec:calibration}.
Samples from the posterior distribution after 50,000 sampling steps are shown in Figure~\ref{fig:posterior_pairs} split into pairwise and marginal representations.

\begin{figure}
    \centering
    \includegraphics[width=\textwidth]{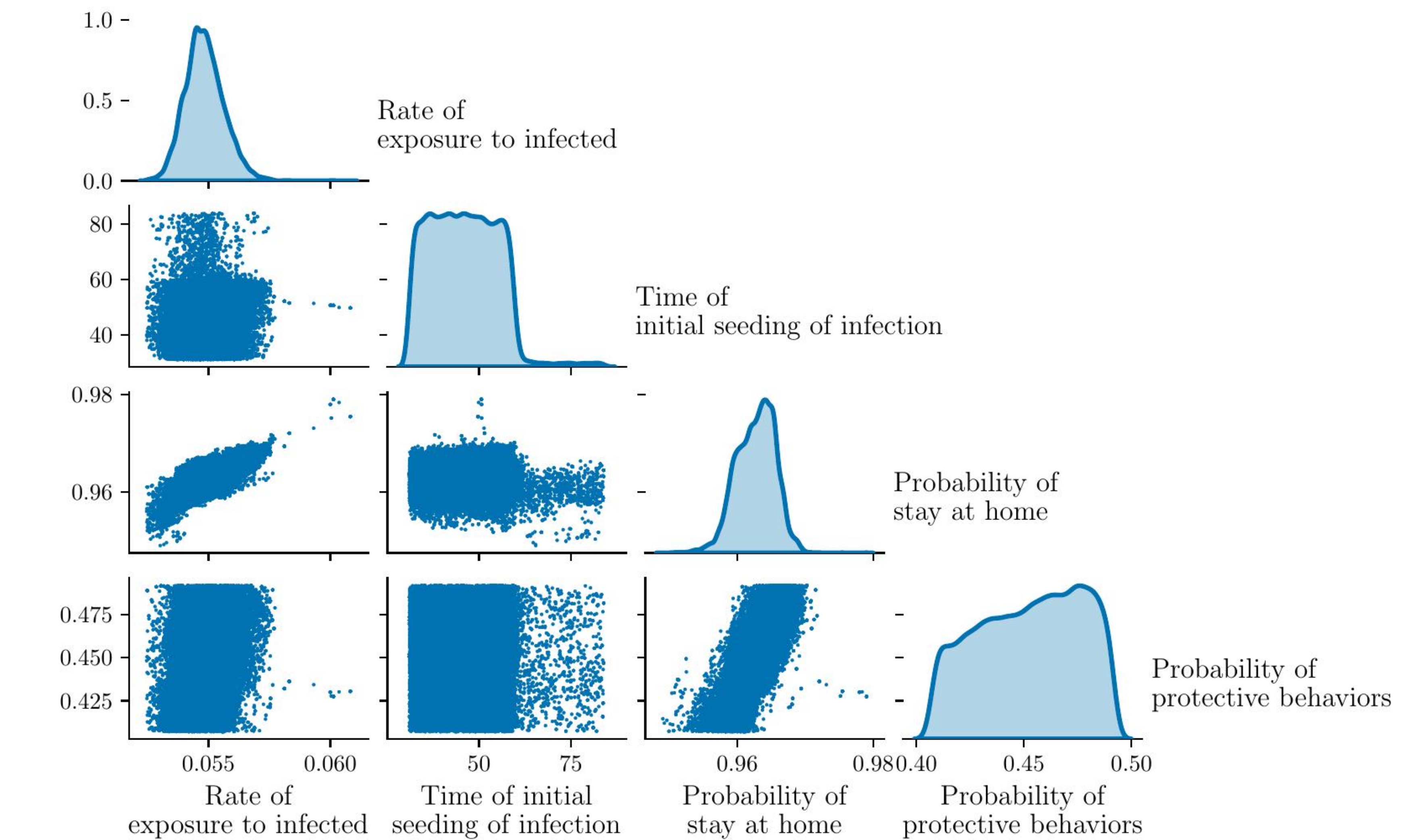}
    \caption{Marginal and pairwise posterior samples from DRAM using the random forest surrogate.}
    \label{fig:posterior_pairs}
\end{figure}

The posterior samples illustrate a more pronounced peak for two of the variables: $\theta_1$ (rate of exposure to infected) and $\theta_2$ (probability of stay at home).
It is also notable that these two parameters show a strong positive correlation.
Namely, if one of the probabilities is increased in CityCOVID, the other must also be increased in order to reasonably match the data.
This result aligns with our understanding of CityCOVID as well as with epidemiological systems.
It can also be observed that the approximated posterior distribution for $\theta_4$ (probability of protective behaviors) is almost uniform in shape.
Though this may be true for CityCOVID as well, it aligns closely with the sensitivities of the random forest shown in Table~\ref{tab:feature_importance}.
Specifically, the lack of importance of this parameter to the random forest allows for almost uniform sampling of its value without significant impactlto the model outputs.

The marginal posterior distributions of this surrogate-based calibration alongside its prior distribution and the IMABC posterior distribution previously computed for CityCOVID\cite{ozik2021population} are shown in Figure~\ref{fig:posterior}. We see that the posterior distributions are peaked and very different from the corresponding priors, implying a significant gain of information regarding parameter values, post calibration, vis-\`a-vis the prior distribution. For $\theta_4$ (probability of protective behavior) we see that the probability density functions (PDFs) from MCMC and IMABC calibration somewhat agree but for the rest, the PDFs computed by MCMC are sharper than those obtained from IMABC.

\begin{figure}
    \centering
    \includegraphics[width=\textwidth]{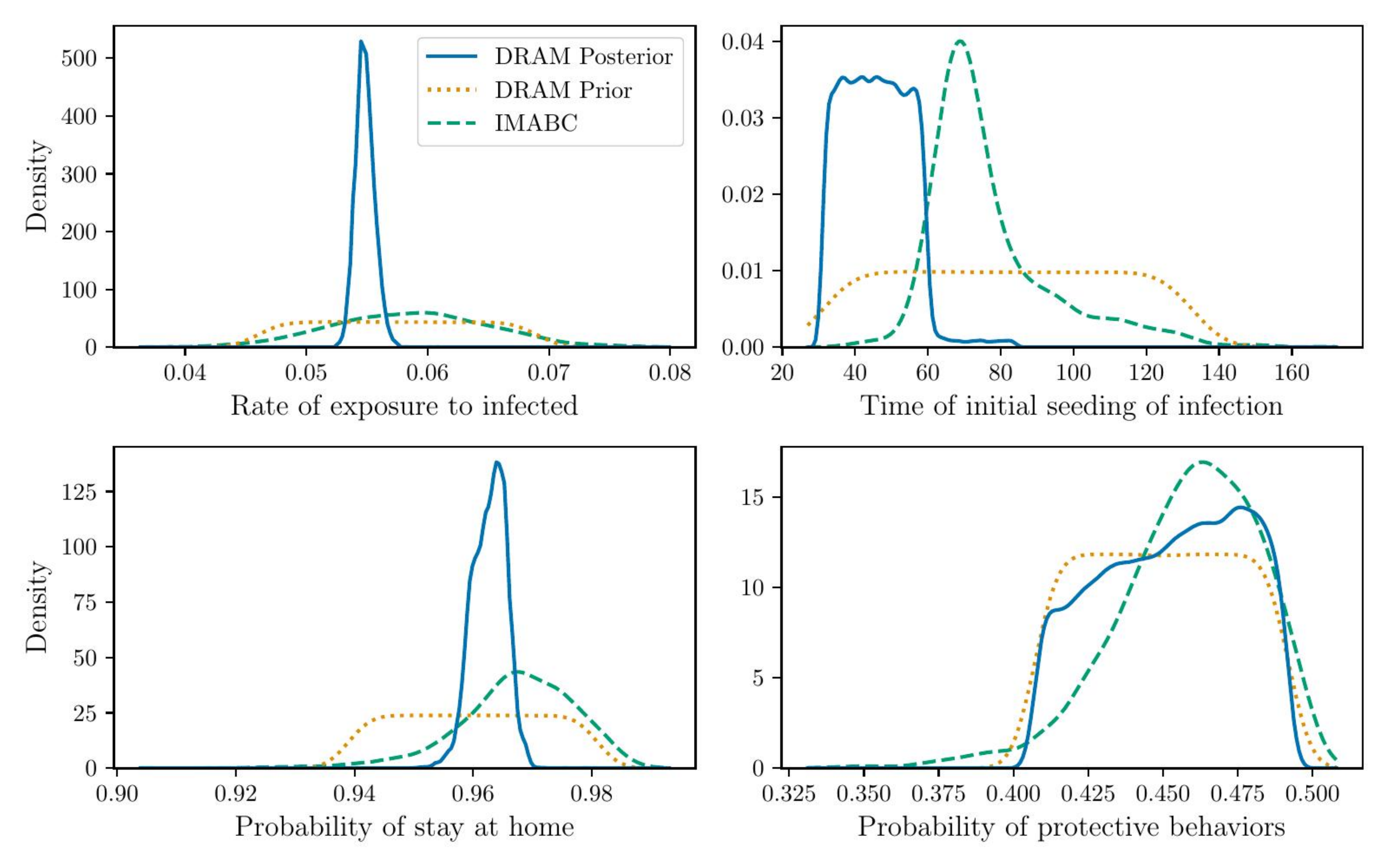}
    \caption{Marginal posterior samples computed with sequential, rejection based IMABC in a previous calibration~\cite{ozik2021population}, from the prior, and from DRAM samples using the random forest surrogate.}
    \label{fig:posterior}
\end{figure}

The posterior distributions are checked via ``pushforwards'' and posterior predictive distributions. In the former, $N_p$ parameters are sampled from the posterior distribution and evaluated (in this case, using the surrogate model) to yield $( \vec{h}, \vec{d})$. In the latter, $\epsilon_h$ and $\epsilon_d$, sampled from their posterior distribution, are added to the pushforward results.
For $N_p = 500$, the trajectories are shown in Figure~\ref{fig:surrogate_predictive}.  The pushforward plots demonstrate that the surrogate is well converged to a narrow band of possible outcomes, which generally follow the trends of daily observed data computed via finite difference. The predictive posterior, which incorporates the calibrated uncertainties $\sigma_h$ and $\sigma_d$ from Equation~\ref{eq:likelihood}, demonstrates almost complete coverage of the observations.
This indicates that the uncertainty in the parameter estimates alone explain very little of the variability of the observations, where those are instead captured with the noise estimates $\sigma_h$ and $\sigma_d$.

\begin{figure}
    \begin{subfigure}{\textwidth}
        \centering
        \includegraphics[width=\textwidth]{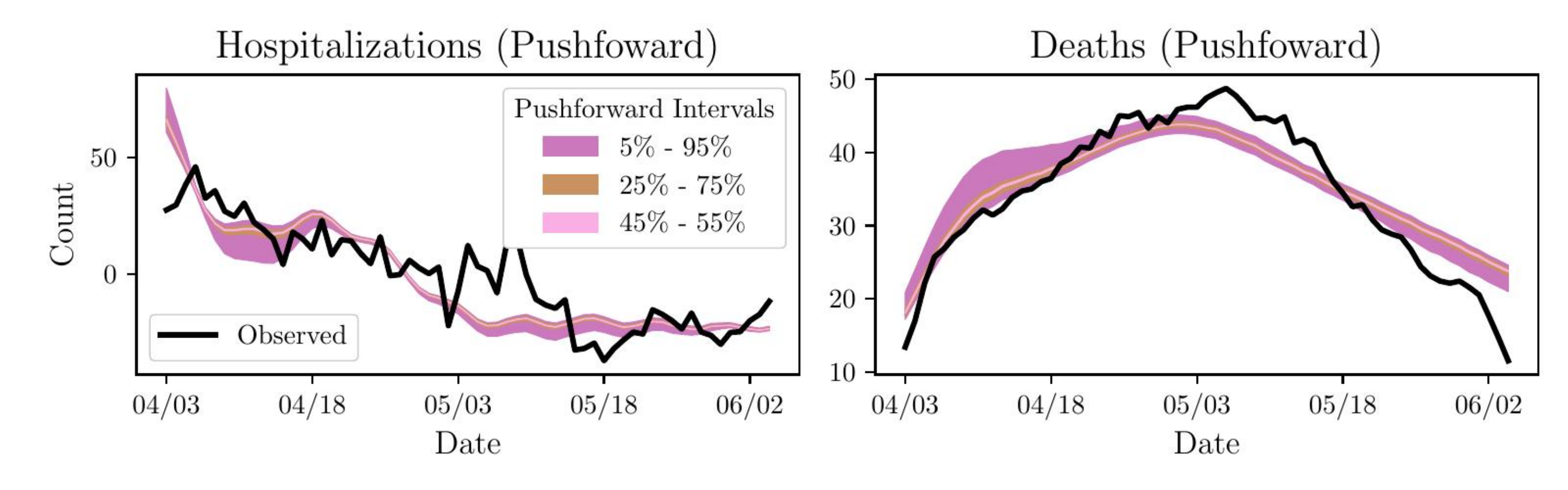}
        \subcaption{}
        \label{fig:posterior_pushforward}
    \end{subfigure}

    \begin{subfigure}{\textwidth}
        \centering
        \includegraphics[width=\textwidth]{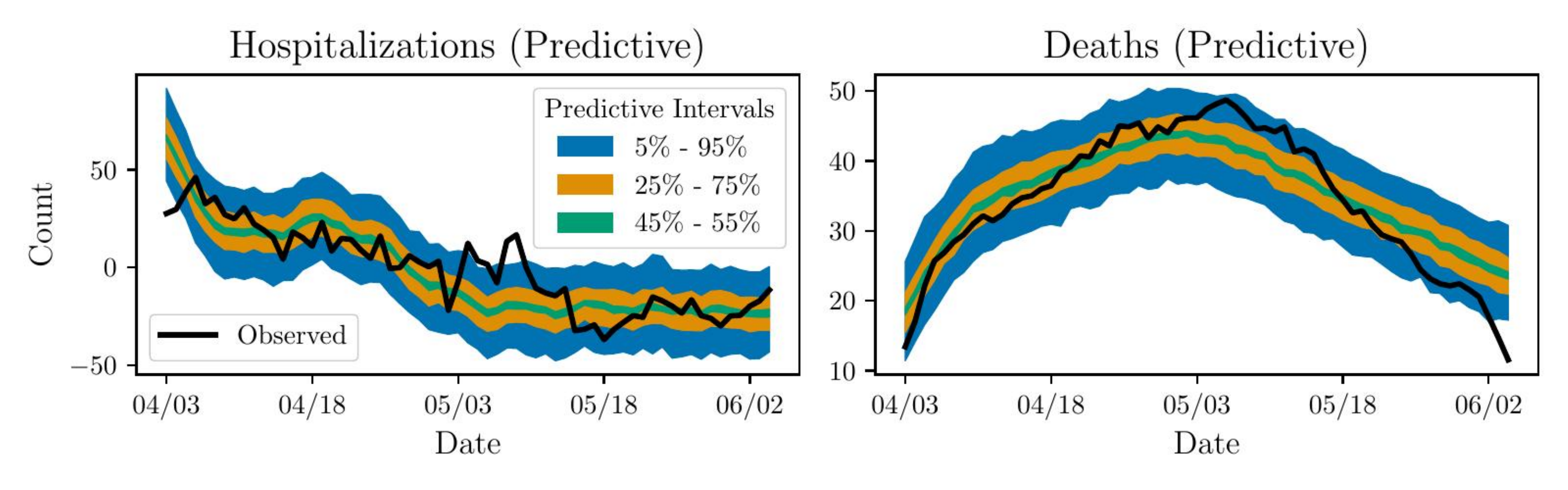}
        \subcaption{}
        \label{fig:posterior_predictive}
    \end{subfigure}

    \begin{subfigure}{\textwidth}
        \centering
        \includegraphics[width=\textwidth]{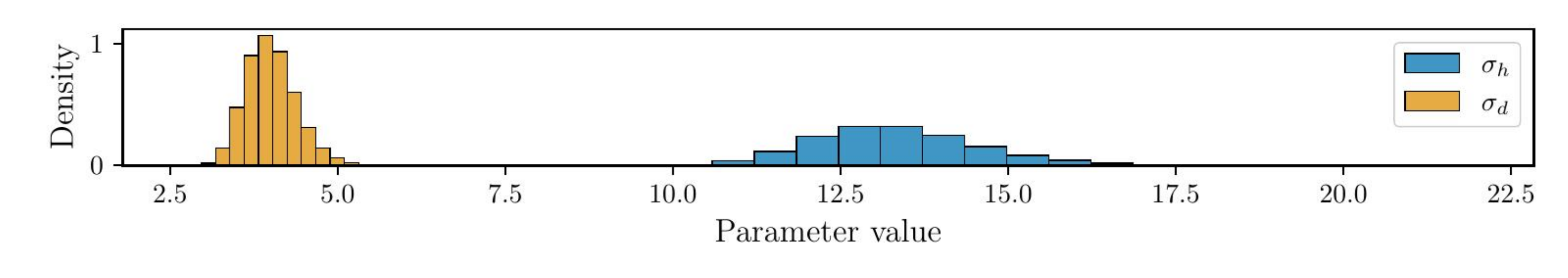}
        \subcaption{}
        \label{fig:posterior_sigmas}
    \end{subfigure}

    \caption{Daily counts of hospitalizations and deaths as observed in Chicago and from the (a) posterior pushforward samples using the surrogate and (b) posterior predictive samples using the surrogate and the calibrated uncertainties (c) $\sigma_h$ and $\sigma_d$ from Equation~\ref{eq:likelihood}. These are obtained using the surrogate model.}
    \label{fig:surrogate_predictive}
\end{figure}

Figure~\ref{fig:rank_verification_hist} plots the VRHs from the surrogate-based calibration (left) and the CityCOVID push-forwards (right, described in more detail in \S~\ref{sec:assess}).
Ideally, this VRH would show a uniform distribution demonstrating that our uncertainty bounds give full and balanced coverage of the true values.
The left subfigure demonstrates that generally the surrogate-based calibration is balanced between over and under prediction (left subfigure).
The VRHs (on the right) from the CityCOVID push-forward runs is far more skewed, showing that while using an approximate surrogate may make the calibration feasible, it incurs an error.

\begin{figure}
    \begin{subfigure}{.49\textwidth}
        \centering
        \includegraphics[width=\textwidth]{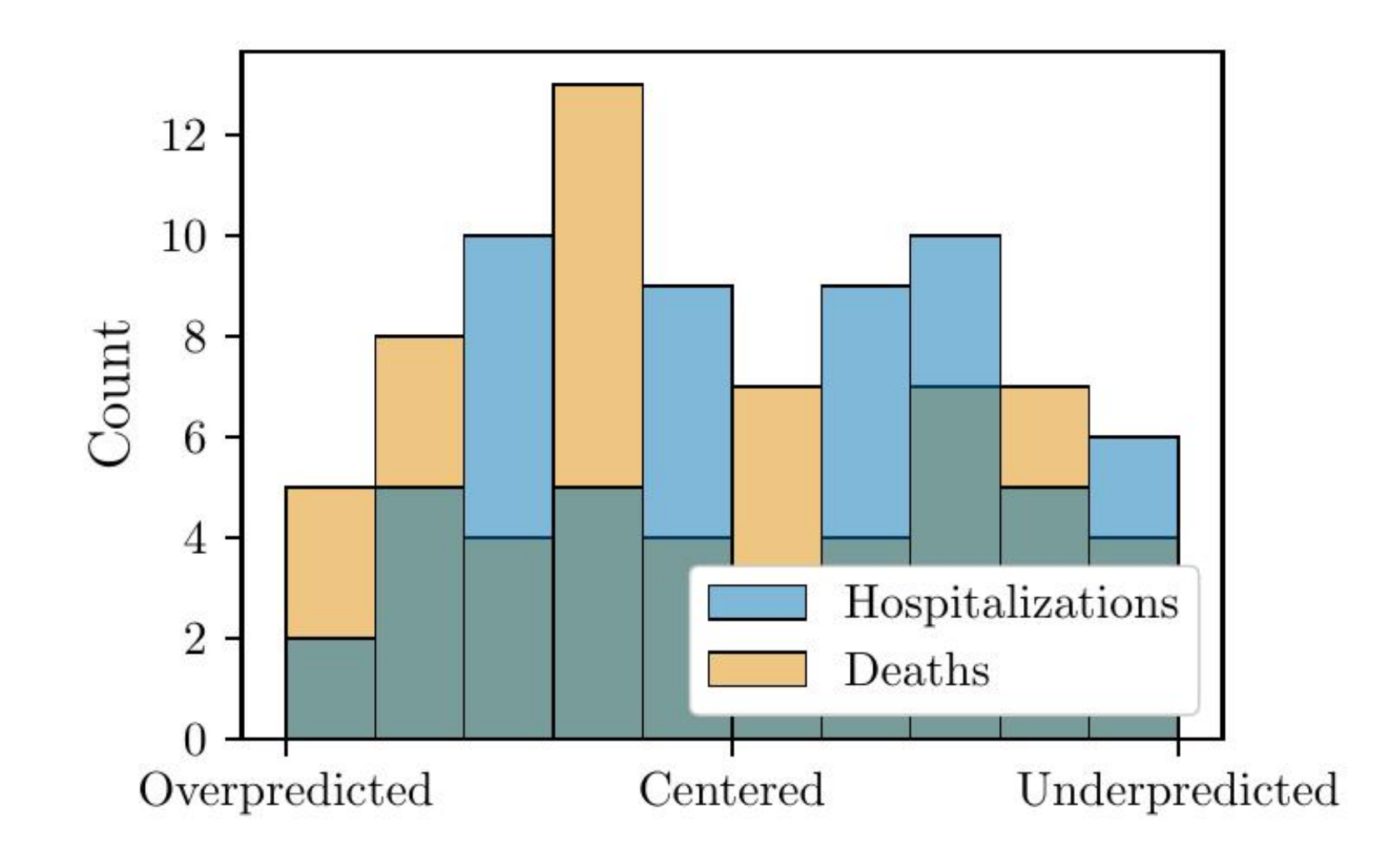}
        \subcaption{}
        \label{fig:posterior_pushforward_rank}
    \end{subfigure}
    \begin{subfigure}{.49\textwidth}
        \centering
        \includegraphics[width=\textwidth]{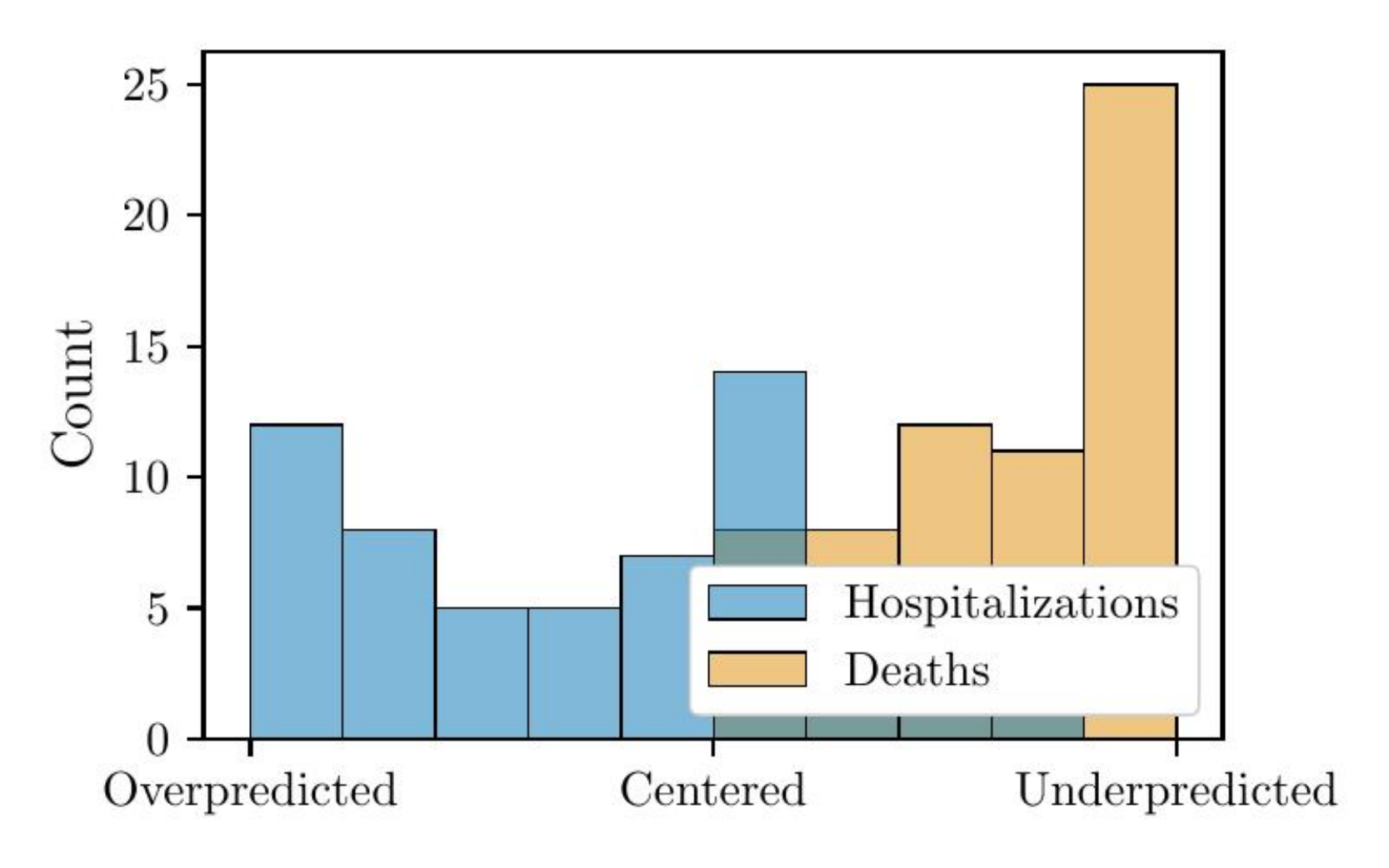}
        \subcaption{}
        \label{fig:posterior_runs_rank}
    \end{subfigure}
    \caption{Verification rank histograms (VRHs) of (a) surrogate and (b) CityCOVID posterior pushforward samples compared with observed data in Chicago. The green color is caused by the overlap of blue and yellow histograms.}
    \label{fig:rank_verification_hist}
\end{figure}

\subsection{Parameter assessment}
\label{sec:assess}

To fully evaluate the quality of the surrogate-based calibration, 100 parameter values were sampled from the approximated posterior distribution and were subsequently run through CityCOVID (using 50 random seeds for each parameter set as was done for the surrogate training set). The resulting posterior pushforward distribution is shown in Figure~\ref{fig:posterior_runs} alongside the pushforward produced in the IMABC calibration~\cite{ozik2021population} in Figure~\ref{fig:abc_runs}.

\begin{figure}
    \begin{subfigure}{\textwidth}
        \centering
        \includegraphics[width=\textwidth]{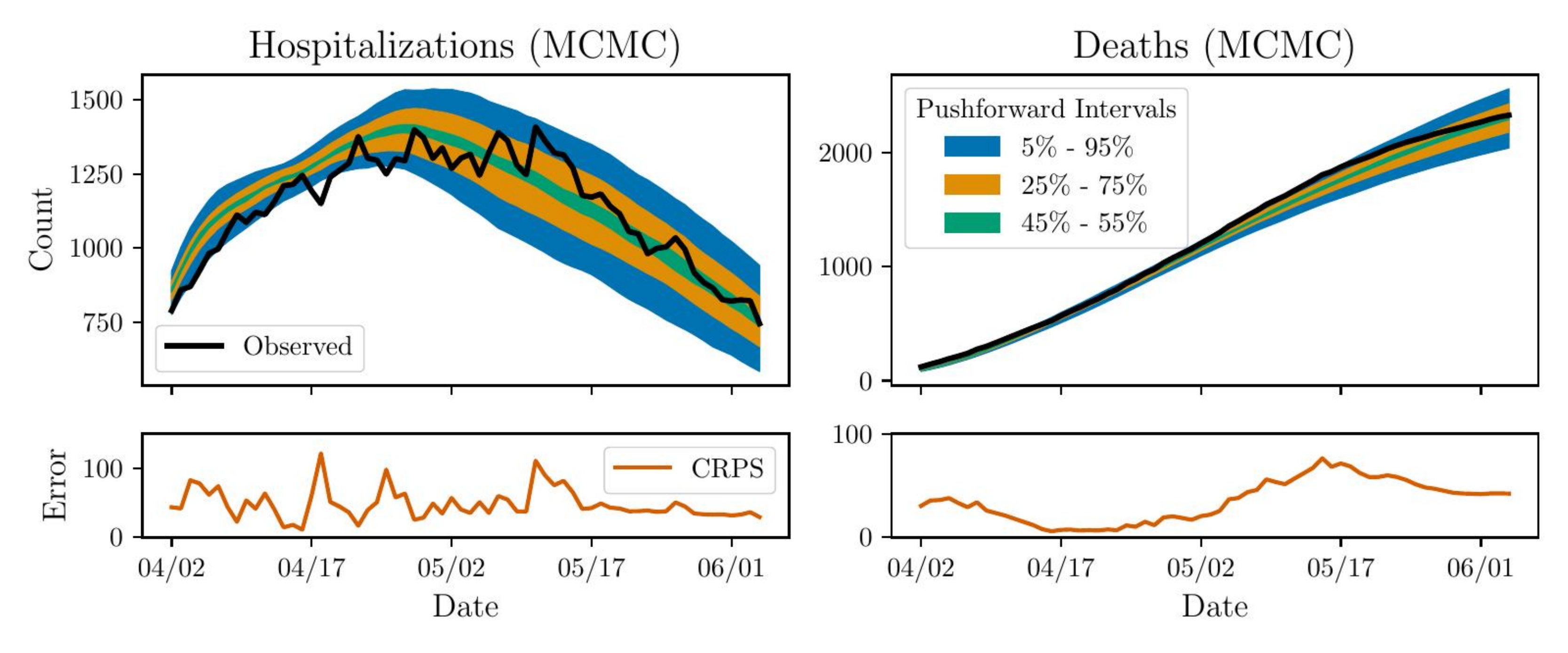}
        \subcaption{}
        \label{fig:posterior_runs}
    \end{subfigure}
    \begin{subfigure}{\textwidth}
        \centering
        \includegraphics[width=\textwidth]{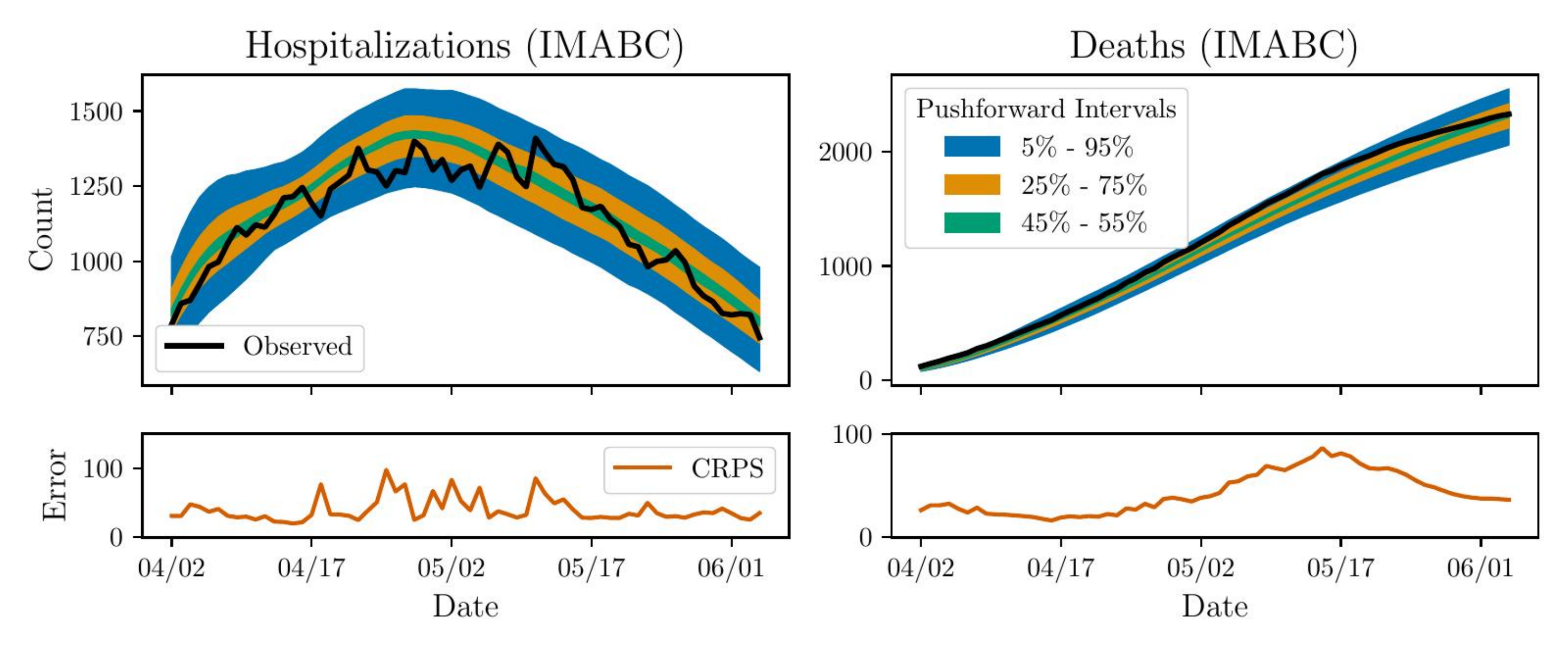}
        \subcaption{}
        \label{fig:abc_runs}
    \end{subfigure}
    \caption{Posterior pushforward runs for samples from posteriors calibrated with (a) DRAM and the random forest surrogate and (b) IMABC sampling as done previously~\cite{ozik2021population}. These results are produced using CityCOVID natively, rather than the surrogate model.}
    \label{fig:pushforward_runs}
\end{figure}

These distributions demonstrate the accuracy of the DRAM and surrogate method when compared to the native IMABC calibration.
Though significantly more efficient in computation, the surrogate approach produced similar results.
It can be seen that the surrogate-based calibration somewhat over predicted during early times and did not capture the full uncertainty.
This can be more precisely observed by considering a proper scoring rule such as the continuous rank probability score (CRPS) which is shown in brown in Figure~\ref{fig:pushforward_runs} and was calculated with the \texttt{scoringutils} R library~\cite{bosse2022scoringutils}. In hospitalizations, the surrogate-based calibration is seen to be less accurate for early times, but is roughly equivalent for the remainder. Alternatively, the deaths from the surrogate-based calibration are slightly more accurate for late times.

Another consideration in this comparison is the disparate number of parameters used for each calibration. Specifically, the IMABC calibration made use of the 9 parameters in Table~\ref{tab:citycovid-details} while the surrogate-based calibration presented here used only the 4 marked with stars.
To compare these disparate measures, the deviance information criterion (DIC~\cite{Gelman:2003}) was used to measure the distance between each posterior pushforward and the observed data while taking into account the number of parameters.
This metric is a generalization of the Akaike Information Criterion (AIC) for Bayesian model comparisons and can be written as:
\begin{align}
    \text{DIC} &= -2 \log p(y \mid \hat{\theta}) + 2p_{\text{DIC}}, \nonumber \\
    p_\text{DIC} &= 2 \left(\log p(y \mid \hat{\theta}) + \mathbb{E}_\text{post}\log p(y \mid \theta)\right).
    \label{eq:dic}
\end{align}
DIC balances the accuracy of the Bayes estimate $\hat{\theta}$ and the effective number of parameters $p_\text{DIC}$.
Given the empirical distribution of posterior samples used for our pushforward distribution, the DIC was estimated with sample means:
\begin{align}
    \hat{\theta} &\approx \frac{1}{N_p} \sum_{i=1}^{N_p} \theta_i, \nonumber \\
    \mathbb{E}_\text{post}\log p(y \mid \theta) &\approx \frac{1}{N_p}\sum_{i=1}^{N_p} \log p(y \mid \theta_i),
    \label{eq:dic_sample}
\end{align}
where $\theta_i$ is the $i^{\text{th}}$ sample from the calibrated posterior distribution used to compute the pushforward distribution.
Accordingly, for the MCMC calibrated posterior $N_p = 100$ and for the IMABC calibrated posterior $N_p = 1158$.

Comparing the DIC calculated for the MCMC and IMABC calibrated pushforward ensembles shows that although the CRPS of the MCMC pushforward higher than the IMABC pushforward, the predictive accuracy of the two approaches was close after accounting for the number of effective parameters.
The numerical comparisons of these quantities averaged over time can be seen in Table~\ref{tab:calibration_comparison}.
The DIC, which penalizes overly parlmetrized models, indicates in favor of the surrogate-based calibration.

An additional layer of comparison can be achieved via comparison of the VRHs of the IMABC and MCMC calibrated pushforwards.
It is agreed that a uniform distribution is ideal for an ensemble forecast~\cite{gneiting2007probabilistic}.
We thus constructed density normalized empirical distributions for the VRHs of the IMABC and MCMC pushforward trajectories and compared them with the corresponding discrete uniform distribution using KL divergence, Chi-squared distance, and Wasserstein distance.
These measurements each provide a unique comparison between the empirical VRH ($V$) and the corresponding discrete uniform distribution ($U$).
Specifically, measurements can be written as:
\begin{align}
    D_{\text{KL}}(V, U) &= \sum_{x\in \mathcal{X}} V(x) \log\left(\frac{V(x)}{U(x)}\right) \label{eq:kl}\\
    D_{\chi^2}(V,U) &= \sum_{x \in \mathcal{X}} \frac{\left(V(x) - U(x)\right)^2}{U(x)} \label{eq:chisq}\\
    D_{\text{wass}}(V,U) &= \frac{\sum_{x_1 \in \mathcal{X}_1} \sum_{x_2 \in \mathcal{X}_2} \left(V(x_1) - U(x_2)\right)\|x_1 - x_2\|}
    {\sum_{x_1 \in \mathcal{X}_1} \sum_{x_2 \in \mathcal{X}_2} (V(x_1) - U(x_2))} \label{eq:wass}
\end{align}
where $\mathcal{X}$ is the space of bins in our histogram and $\mathcal{X}_1$ and $\mathcal{X}_2$ encode the optimal paths to move mass from $V$ to $U$ (computed via linear optimization).
These can be simply interpreted as follows:
\begin{description}
    \item[KL Divergence (Eq.~\ref{eq:kl})]
    Measurement of the lost information from using $V$ in place of $U$.
    \item[Chi-Sq Distance (Eq.~\ref{eq:chisq})]
    Measurement of the difference in frequencies in the histograms.
    \item[Wasserstein Distance (Eq.~\ref{eq:wass})]
    Measurement of the histogram density to be moved to align the VRH with the discrete uniform.
\end{description}

All three measures show that the surrogate-based calibration provides a VRH that is closer to a uniform distributiln than the one arising from IMABC.
The tabulated results illustrate the competing effects of approximations in ABM calibration. While IMABC resulted in overly-wide marginal PDFs (see Figure~\ref{fig:posterior}), the surrogate-based calibration is not without its flaws.
While the comparison was not evident pictorially in Figure~\ref{fig:pushforward_runs}, the tabulated summary in Table~\ref{tab:calibration_comparison} shows that the MCMC calibration is slightly superior, despite the use of surrogate models.

However, for both calibrations, the verification rank histogram of these pushforward results seen in Figure~\ref{fig:rank_verification_hist} illustrates a general overprediction for hospitalizations and underprediction for deaths.
In fact, the number of deaths is always underpredicted showing an imbalanced calibration.
This holds true for both the calibrations, which were conducted using independent formulations of the estimation problem as well as using different algorithms.
This error indicates a model-form error in CityCOVID that causes hospitalized people to die at a rate lower than what is observed.


\begin{table}
\centering
\begin{tabular}{|l|c|c|c|c|}
\hline
\rowcolor[gray]{0.8}
&
\multicolumn{2}{c|}{\textbf{Hospitalizations}} &
\multicolumn{2}{c|}{\textbf{Deaths}}
\\ \hhline{|*1{>{\arrayrulecolor[gray]{.8}}-}*4{>{\arrayrulecolor{black}}-}|}
\rowcolor[gray]{0.8}
\multirow{-2}{*}{\textbf{Metric} (lower values are better)} &
\textbf{IMABC} &
\textbf{MCMC} &
\textbf{IMABC} &
\textbf{MCMC}

\\ \hline
\textbf{CRPS} &
39.74 &
47.85 &
42.53 &
34.96
\\ \hline
\textbf{DIC} &
685.79 &
596.50 &
16.40 &
9.95
\\ \hline
\textbf{KL Divergence (VRH)} &
0.33 &
0.21 &
0.88 &
0.79
\\ \hline
\textbf{Chi-Sq Distance (VRH)} &
41.00 &
24.13 &
114.75 &
95.06
\\ \hline
\textbf{Wasserstein Distance (VRH)} &
0.07 &
0.05 &
0.11 &
0.10
\\ \hline
\end{tabular}
\caption{Comparison of pushforward results of samples from surrogate-based MCMC calibration with previous approximate Bayesian calibration~\cite{ozik2021population}. Rows show the continuous rank probability scores (CRPS), the deviance information criterion (DIC), and the KL-divergence, Chi-Squared distance, and Wasserstein distance of the verification rank histogram (VRH) from a uniform distribution.}
\label{tab:calibration_comparison}
\end{table}


\section{Conclusion}

We have described an accelerated approach to calibrate agent-based models for epidemiology in which the quantities of interest are population level metrics such a hospitalizations and deaths.
In order to overcome the inherent stochasticity of these models, we considered a mean model which was averaged over random seeds.
The temporal dynamics of the quantities of interest were then decomposed via PCA and a random forest was trained to reconstruct the data using these principal components and the input parameters for the ABM.
This combination yielded a surrogate which could be used in place of the model for accelerated sampling.

In order to effectively use this surrogate on the model of interest, the Gini impurity of the random forest trained on some preliminary model outputs was used to reduce the parameter space to only 4 dimensions.
A hypercube was then sampled in this lower dimensional space to provide training information for the surrogate.
An empirical prior was then constructed which combined samples of the hypercube yielding hospitalization and death trajectories near those observed.

Equipped with the surrogate model and an adequate empirical prior, Markov chain Monte Carlo sampling was used with a Gaussian error model to sample from the posterior distribution.
The samples, along with their respective pushforward and posterior predictive trajectories, were analyzed in comparison with a previous IMABC calibration of the model\cite{ozik2021population}.
Ultimately, this surrogate accelerated approach yielded similar results to the original approach at a fraction of the computational cost.
True posterior pushforward samples in combination with verification rank histograms and proper scoring rules such as CRPS demonstrated that the loss in accuracy of the accelerated surrogate-based calibration was almost negligible.

However, the final calibration using either IMABC or the full Bayesian inference approach tend to over or underpredict the values of interest.
Future work will aim to correct this inaccuracy by doing a more complete incorporation of the stochasticity of the model.
Several approaches have been proposed for this more complete analysis including fitting surrogates to approximate both the mean/median output across random seeds as well as the variance/quantiles across the random seeds~\cite{fadikar2018abm_surrogates_gp_calibration} and fitting surrogates using the random seeds themselves~\cite{fadikar_trajectory-oriented_2023}.

Additionally, the surrogate used for this calibration was constructed in a global fashion on few dimensions and its limits are not yet well understood.
We plan future work to compare this construction with alternative calibration approaches which can scale to larger dimensional problems or for which local surrogate construction can be combined with native model outputs to reduce the impact of surrogate model form error.

\clearpage
\section*{\bf Author contributions}

Connor Robertson formulated the problem, wrote the software to solve it, generated the figures and wrote the paper. Cosmin Safta assisted with software development, interpretation of results, and contributed to writing the paper. Nicholson Collier produced the CityCOVID training data. Jonathan Ozik provided guidance on CityCOVID and IMABC calibration and contributed to writing the paper. Jaideep Ray posed the problem, assisted with the epidemiological interpretation, and suggested the calibration approach and metrics.

\section*{Acknowledgments}
We thank Arindam Fadikar and Chick Macal at Argonne National Laboratories for various useful discussions on surrogates and their application to epidemiological modeling.
This paper describes objective technical results and analysis. Any subjective views or opinions that might be expressed in the paper do not necessarily represent the views of the U.S. Department of Energy or the United States Government. This article has been authored by an employee of National Technology \& Engineering Solutions of Sandia, LLC under Contract No. DE-NA0003525 with the U.S. Department of Energy (DOE). The employee owns all right, title and interest in and to the article and is solely responsible for its contents. The United States Government retains and the publisher, by accepting the article for publication, acknowledges that the United States Government retains a non-exclusive, paid-up, irrevocable, world-wide license to publish or reproduce the published form of this article or allow others to do so, for United States Government purposes. The DOE will provide public access to these results of federally sponsored research in accordance with the DOE Public Access Plan https://www.energy.gov/downloads/doe-public-access-plan.  This material is based upon work supported by the National Science
Foundation under Grant 2200234, the U.S. Department of Energy, Office of Science, under contract number DE-AC02-06CH11357 and the Bio-preparedness Research Virtual Environment (BRaVE) initiative. This research was completed with resources provided by the Laboratory Computing Resource Center at Argonne National Laboratory.

\section*{\bf Financial disclosure}
None reported.

\section*{\bf Conflict of interest}
The authors declare no potential conflict of interests.

\bibliographystyle{plain}
\bibliography{references}

\newpage

\appendix

Code to reproduce results from this article can be found at \url{https://github.com/sandialabs/Bayesian-calibration-of-stochastic-agent-based-model-via-random-forest}.

\section{Surrogate performance} \label{asec:surrogate_performance}

The random forest surrogate has no formal guarantees to match CityCOVID.
As a result, its sensitivity and accuracy need to be independently verified.
The sensitivity of the random forest to the most impactful ABM parameters is shown in Table~\ref{tab:rf_sensitivity} for various sensitivity measures:
\begin{description}
    \item[Gini importance]
    A measure of the frequency with which a parameter is used for splits within the trees of the forest.
    More frequent splitting is indicative of the forest's reliance on information from that parameter.
    \item[Permutation importance]
    A measure of accuracy of the forest when the input data of a parameter is shuffled.
    Severely reduced accuracy when a single parameter is shuffled indicates the forest's reliance on information from that parameter.
    \item[Sobol (first)]
    A measure of the variance of the output of the random forest across variations in a parameter.
    Significant changes in output from adjustments to a single parameter indicates the forest's reliance on information from that parameter.
    \item[Sobol (total)]
    A measure of the variance of the output of the random forest across variations in a parameter and that parameter in combination with others.
    Significant changes in output from adjustments to a single parameter indicates the forest's reliance on information from that parameter.
    This total form also attempts to include nonlinear interactions with other parameters.
\end{description}
\begin{table}[!h]
\centering
\begin{tabular}{|l|c|c|c|c|}
\hline
\rowcolor[gray]{0.8}
\textbf{Feature} &
\textbf{Gini Importance} &
\textbf{Permutation Importance} &
\textbf{Sobol (first)} &
\textbf{Sobol (total)}
\\ \hline
\bf{Rate of exposure to infected} &
0.17 &
0.06 &
0.05 &
0.09
\\ \hline
\bf{Time of initial exposure} &
0.17 &
0.24 &
0.26 &
0.37
\\ \hline
\bf{Probability of stay at home} &
0.11 &
0.16 &
0.34 &
0.44
\\ \hline
\bf{Probability of protective behaviors} &
0.11 &
0.04 &
0.04 &
0.07
\\ \hline
Shielding by other susceptible&
0.10 &
0.03 &
0.05 &
0.08
\\ \hline
Number of initially infected &
0.09 &
0.03 &
0.03 &
0.06
\\ \hline
Seasonality multiplier &
0.08 &
0.03 &
0.01 &
0.02
\\ \hline
Proportion isolating in nursing home &
0.07 &
0.01 &
0.01 &
0.01
\\ \hline
Proportion isolating in home &
0.07 &
0.01 &
0.00 &
0.01
\\ \hline
\end{tabular}
\caption{Random forest feature importance metrics using 9 identified important parameters and data from IMABC calibration~\cite{ozik2021population}. Parameters selected by analysis of the surrogate sensitivity are bolded.}
\label{tab:rf_sensitivity}
\end{table}

After reducing the input parameters of the random forest surrogate, accuracy of the surrogate can be determined via the median absolute relative error as shown in Figure~\ref{fig:surrogate_err}.
Some concrete examples demonstrating the absolute relative error for several output trajectories can be seen in Figure~\ref{fig:surrogate_examples}.
These examples visually demonstrate the large relative errors for small values of hospitalizations and deaths which encouraged the use of the median as a measure of accuracy.

\begin{figure}
    \begin{subfigure}{\textwidth}
      \includegraphics[width=\textwidth]{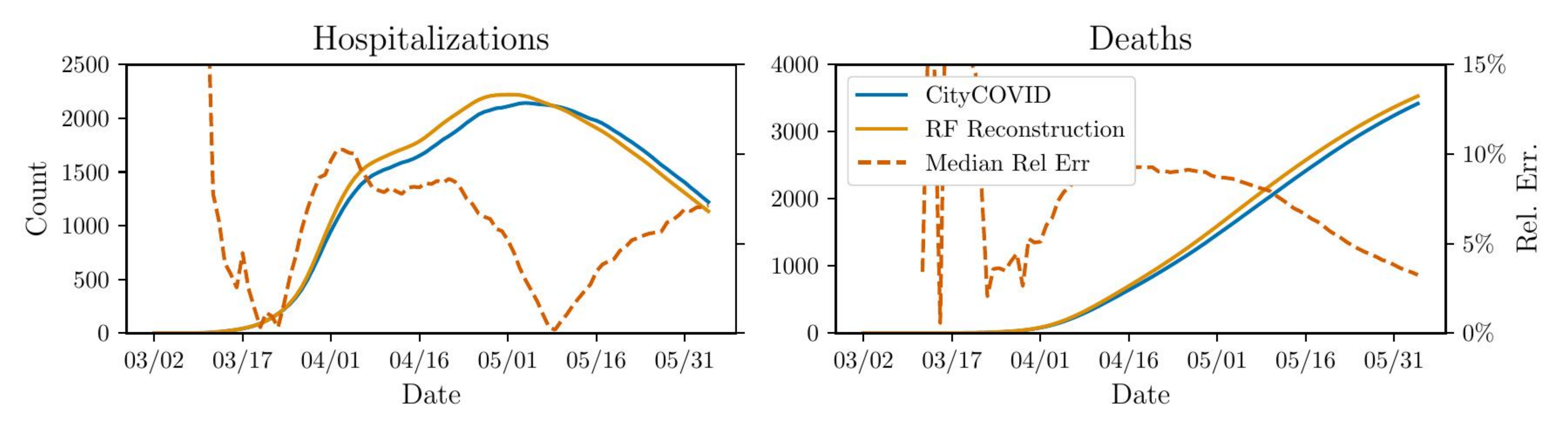}
      \subcaption{}
      \label{fig:rf_reconstruct_32}
    \end{subfigure}
    \begin{subfigure}{\textwidth}
      \includegraphics[width=\textwidth]{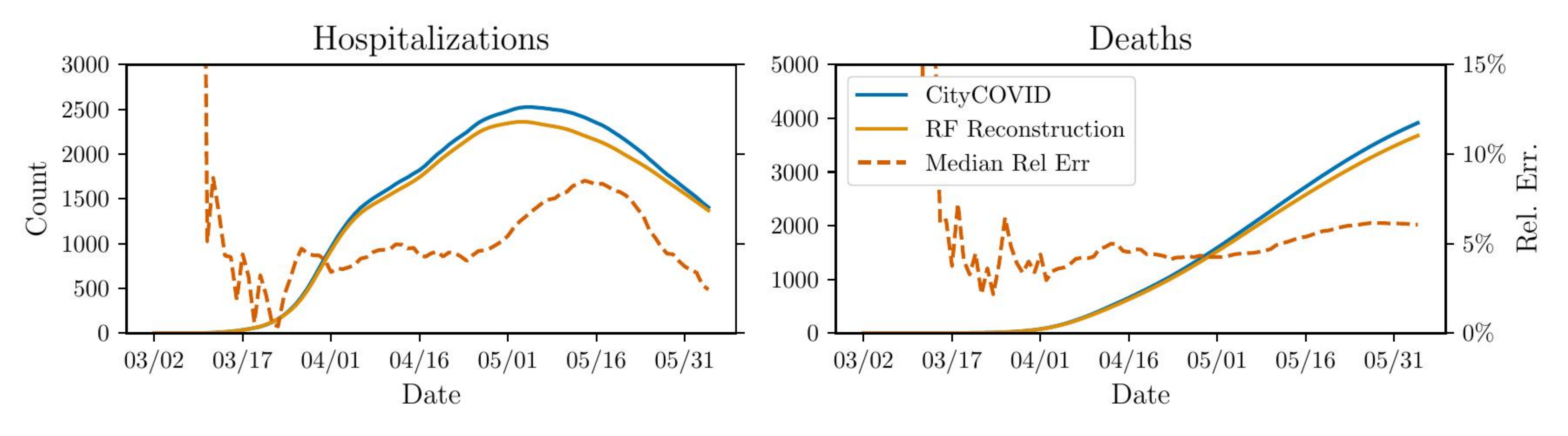}
      \subcaption{}
      \label{fig:rf_reconstruct_481}
    \end{subfigure}
    \begin{subfigure}{\textwidth}
      \includegraphics[width=\textwidth]{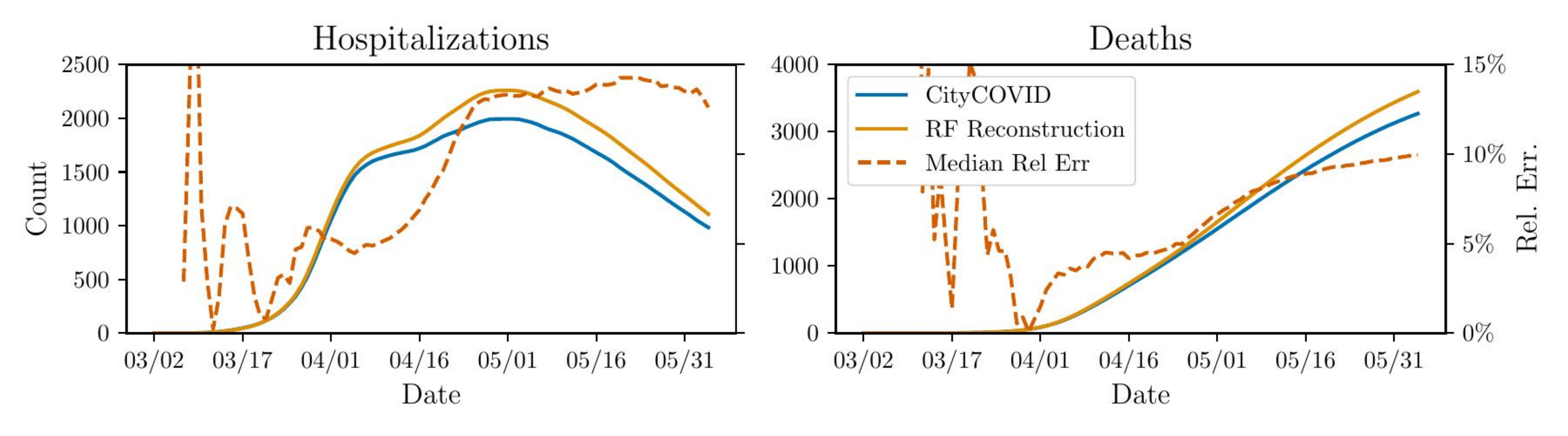}
      \subcaption{}
      \label{fig:rf_reconstruct_627}
    \end{subfigure}
    \caption{Reconstruction of the data using the surrogate for several different runs of CityCOVID. The relative error can be observed to be significantly larger for small values of hospitalizations and deaths. Here ``RF Reconstruction'' implies predictions with our surrogate model and ``Rel. Err'' implies Relative Error.}
    \label{fig:surrogate_examples}
\end{figure}

\end{document}